\documentclass[letterpaper, 10 pt, conference]{ieeeconf} 

\IEEEoverridecommandlockouts               
 
\usepackage[pdftex]{graphicx}
\usepackage{amssymb}
\usepackage{amsmath}
\usepackage{bm}
\usepackage{subfigure}
\usepackage{cite}
\usepackage{epsfig}
\usepackage{color}
\usepackage{psfrag}
\usepackage{cases}
\usepackage{acronym}
\usepackage{latexsym}
\usepackage{dblfloatfix}

\title{\LARGE \bf
Robust Visual SLAM with Point and Line Features
}

\author{Xingxing Zuo$^{1}$, Xiaojia Xie$^{1}$, Yong Liu$^{1,2}$ and Guoquan Huang$^{3}$
	\thanks{$^{1}$Xingxing Zuo and Xiaojia Xie are with the Institute of Cyber-Systems and Control, Zhejiang University, Zhejiang, 310027, China.}%
	\thanks{$^{2}$Yong Liu is with the State Key Laboratory of Industrial Control Technology and Institute of Cyber-Systems and Control, Zhejiang University, Zhejiang, 310027, China (Yong Liu is the corresponding author, email: yongliu@iipc.zju.edu.cn).}%
	\thanks{$^{3}$Guoquan Huang is with the Department of Mechanical Engineering, University of Delaware, Newark, DE 19716, USA.}%
}

\begin{document}

\maketitle
\thispagestyle{empty}
\pagestyle{empty}

\begin{abstract}

In this paper, we develop a robust efficient visual SLAM system that utilizes heterogeneous point and line features. By leveraging ORB-SLAM~\cite{murORB2}, the proposed system consists of stereo matching, frame tracking, local mapping, loop detection, and bundle adjustment of both point and line features. In particular, as the main theoretical contributions of this paper, we, for the first time, employ the orthonormal representation as the minimal parameterization to model line features along with point features in visual SLAM and analytically derive the Jacobians of the re-projection errors with respect to the line parameters, which significantly improves the SLAM solution. The proposed SLAM has been extensively tested in both synthetic and real-world experiments whose results demonstrate that the proposed system outperforms the state-of-the-art methods in various scenarios.

\end{abstract}

\section{Introduction}

Visual SLAM (V-SLAM) is one of enabling technologies for autonomous systems such as self-driving cars, unmanned aerial vehicles and space robots.
While most V-SLAM solutions rely on point features due to their simplicity,
line features commonly seen in man-made environments
are less sensitive to lighting variation and position ambiguity and have been only used in recent work~\cite{rother2003linear,marzorati2007integration,klein2008improving,koletschka2014mevo,zhang2015building,lu2015robust,gomez2016robust}. 
In principle, the combination of point and line features would provide more geometric constraints about the structure of the environment than either one, 
which motivates us to design robust V-SLAM with point and line features.

 
 Recently, optimization-based approaches have become favorable for the V-SLAM due to its superior accuracy per computational unit as compared with filtering-based approaches~\cite{strasdat2010real}. 
In particular, graph-based SLAM is one of the most popular formulations which constructs a factor graph whose nodes correspond to the states to estimate and edges represent measurement constraints between the nodes.
%
When incorporating the line features into the traditional point feature-based graph SLAM framework, two challenges arise:
The first one is that the spatial line is often over parameterized for the convenience of transformation~\cite{lu2015robust,klein2008improving,marzorati2007integration}, 
which incurs extra computational overhead in the graph optimization. 
{Note that while a spatial line has only {\em four} degrees of freedom, typically it is represented by its two spatial endpoints or the $ Pl\ddot{u}cker $ coordinates with {\em six} degrees of freedom.}
Secondly, 
it is known that the Jacobian plays an important role when using an iterative approach to solve the graph optimization problem.
In part because of the over parametrization, most approaches~\cite{lu2015robust,zhang2015building} using line features typically employ the numerically computed Jacobians, which incurs approximation. 
In contrast, we analytically compute the Jacobians during the graph optimization in order to improve accuracy as well as efficiency.


 
 In particular, this paper introduces a robust and efficient graph-based visual SLAM system using both point and line features with a unified cost function, 
combining the re-projection errors of points and lines. 
In our back-end, the spatial line is parametrized by the orthonormal representation, which is the minimal and decoupled representation. 
Based on this minimal parametrization, we further derive the analytical Jacobian of the line re-projection error.
Specifically, the main contributions of this paper are the following:
\begin{itemize}
	\item An improved extraction and matching method for line features is introduced to robustify data association. 
	
	\item	In the back-end of the proposed visual SLAM, we employ the orthonormal (minimal) representation to parameterize lines and analytically compute the corresponding Jacobians. 
	
		\item	We design and implement a complete visual SLAM system using both point and line features, which includes stereo matching, frame tracking, local mapping, bundle adjustment of both line feature and point feature, as well as point-line based loop detection. Extensive experimental results are presented to validate its performance.
\end{itemize}

\begin{figure*}[!t]
	\centering
	\subfigure [Point and line features detected in one image]{\includegraphics[scale=.2]{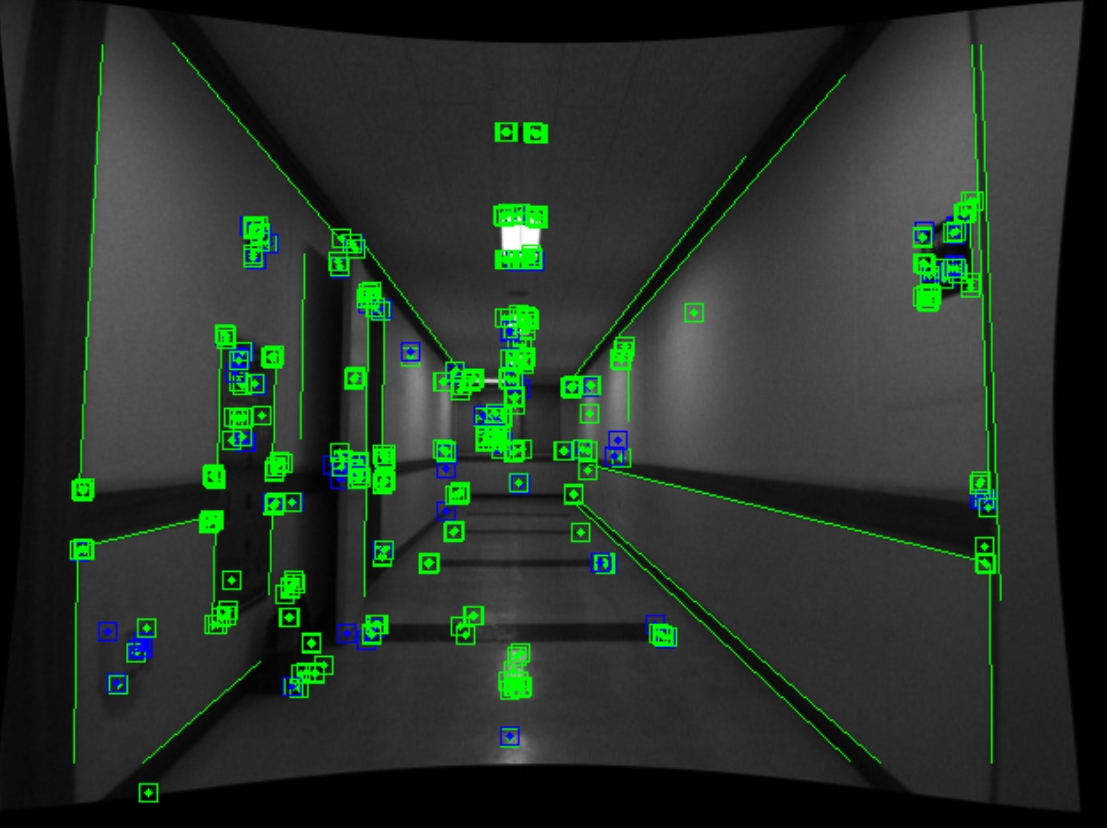}}~~
	\subfigure	[The point and line map]{\includegraphics[scale=.3]{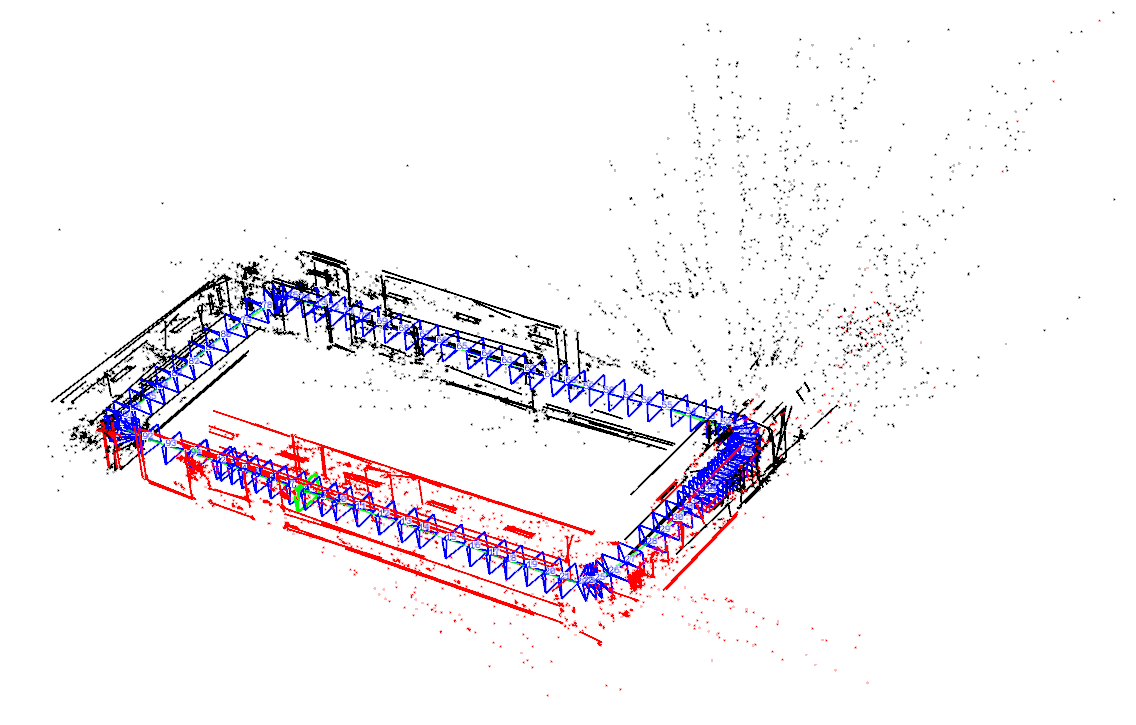}}
	
	\caption{The proposed visual SLAM with point and line features on the it3f dataset \cite{zhang2015building}. 
	Note that in (b), the green lines indicate the trajectory of camera motion. The blue frames represent keyframes, the current frame in green, and the local map for the tracking at that moment in red. }
	\label{ifig1}
\end{figure*}



\section{Related Work}

 Some methods have been proposed to parameterize line in three-dimensional (3D) space efficiently. Sola \cite{sola2012impact} summarizes several methods to represent line including $ Pl\ddot{u}cker $ coordinates, Anchored $ Pl\ddot{u}cker $ Lines, and homogeneous-points line etc. For minimizing the number of the parameters, Bartoli \cite{bartoli2005structure} proposed the orthonormal representation with the minimum four parameters to represent spatial lines in SFM. 

Combination of point and line features has been utilized in the SLAM community recently. Marzorati et al.~\cite{marzorati2007integration} proposed a SLAM with points and lines, which uses a special trifocal cameras to detect and reconstruct 3D lines. Rother~\cite{rother2003linear} reconstructed points and lines at the cost of requiring a reference plane to be visible in all views. Koletschka et al.~\cite{koletschka2014mevo} proposed a stereo odometry based on points and lines, which computes the sub-pixel disparity of the endpoints of the line and deals with partial occlusions. Lu \cite{lu2015robust} fuses point and line features to form a RGBD visual odometry algorithm, which extracts 3D points and lines from RGB-D data. It is also proved that fusing these two types of features produced smaller uncertainty in motion estimation than using either feature type alone in his work. Ruben \cite{gomez2016robust} proposed a probabilistic approach to stereo visual odometry based on the combination of points and line segments, which weighs the associated errors of points and line segments according to their covariance matrices. 

%

\section{Detection and Representation of Line Features}

\subsection{Extraction and Description of Line Features}

\begin{figure}[!h]
	\centering
	\includegraphics[width=3.2in,height=0.8in ]{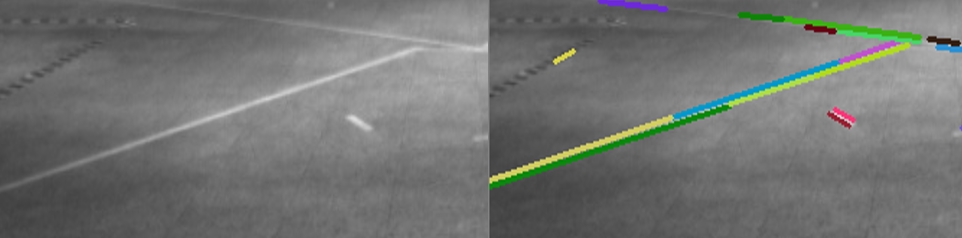}
	\caption{Performance of LSD. Left: Original image. Right: Line features detected in the image by LSD.}
	\label{rfig1}
	\end{figure}

 Line Segment Detector (LSD) \cite{von2010lsd} is a popular feature detector for line segments.
 It is designed to work on noisy image in various scenes without parameter tuning and is able to provide subpixel accuracy. 
However, the LSD suffers from the problem of dividing a line into multiple segments in some scenarios as shown in Fig.~\ref{rfig1},
 causing failures for matching and tracking line features. 

  \begin{figure}
	\centering
	\includegraphics[width=2in,height=0.4in ]{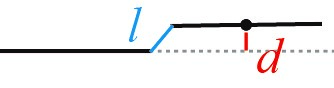}
	\caption{Distances between two line segments.}
	\label{rfig2}
\end{figure}

\begin{figure}[!b]
	\centering
	\includegraphics[width=3.2in,height=0.8in ]{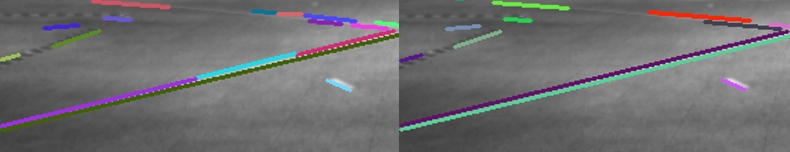}
	\caption{ Comparative results of different detectors. Left: The original LSD detector. Right: The proposed improved detector.}
	\label{rfig3}
\end{figure}

Therefore, 
in this paper, we seek to improve the LSD algorithm by minimizing the influence of dividing a line into multiple segments. 
In particular, we merge the line segments which should be on the same one straight line while being divided to several parts. For each line segment extracted by LSD, the start point and end point can be distinguished, because the direction is encoded by which side of the line segment is darker.
 In our improvement, we merge the segments according to their differences of both direction and distance. As shown in Fig. \ref{rfig2}, $ l $ represents the minimum distance between the endpoints of the two segments, and $ d $ indicates the distance from the midpoint of one segment to the other line segment. If $ d $ and $ l $ are smaller than the given threshold and the direction difference is also small, then the two segments are considered to be the candidates to be combined. This improved line detector has the advantages of making more robust and accurate data association as demonstrated in our experiments. 
 Fig. \ref{rfig3} shows the result of the two different detectors. Note that the merged line segments found by our improved detector is represented by the LBD line descriptor \cite{Zhang2013}, which is a 256-bit vector same as ORB point descriptor \cite{rublee2011orb}. 
 The distance between the two descriptors can be another criterion for fusing two lines.
 

\subsection{Line Feature Matching}\label{seclinematch}

Based on the LBD line segment descriptor, we introduce the geometric properties \cite{woo20092d} of line to perform effective line matching. In our approach, two successfully matched line features $ \bm{l}_1 $ , $ \bm{l}_2 $ need to satisfy the following conditions:
\begin{enumerate}
	\item the angular difference of two matched lines is smaller than a given threshold $ \Phi $;
	\item the length of $ \bm{l}_1 $, $ \|\bm{l}_1\| $ is similar to the length of $ \bm{l}_2 $, $ \|\bm{l}_2\| $: $\frac{{min\left( {{\|\bm{l}_1\|},{\|\bm{l}_2\|}} \right)}}{{max\left( {{\|\bm{l}_1\|},{\|\bm{l}_2\|}} \right)}} > \tau $;
	\item the overlapping length of the two lines is greater than a certain value: $\frac{{{\bm{l}_{overlap}}}}{{\min ({\|\bm{l}_1\|},{\|\bm{l}_2\|})}} > \beta $;
	\item the distance between the two LBD descriptors is less than a certain value.	
\end{enumerate}

\subsection{Geometric Representation}

As 3D line can be initialized by its two spatial points, we assume that their homogeneous coordinates are 
$\bm{X}_1 = {({x_1},{y_1},{z_1},{r_1})^T}$, $\bm{X}_2 = {({x_2},{y_2},{z_2},{r_2})^T}$ respectively, while the inhomogeneous coordinates are represented as ${\bm{\tilde{X}}_1} $, ${\bm{\tilde{X}}_2} $. 
Then $ Pl\ddot{u}cker $ coordinates can be constructed as follows: 

\begin{equation}
	\bm{\mathcal{L}} = \left[ {\begin{array}{*{20}{c}}
	{{{\bm{\tilde{X}}}_1} \times {{\bm{\tilde{X}}}_2}}\\
	{{r_2}{{\bm{\tilde{X}}}_1} - {r_1}{{\bm{\tilde{X}}}_2}}
	\end{array}} \right] = \left[ {\begin{array}{*{20}{c}}
	\bm{n}\\
	\bm{v}
	\end{array}} \right] \in {\mathbb{P}^5} \subset {\mathbb{R}^6}
\label{3eq1}
\end{equation}
which is a 6-dimensional vector consisting of $ \bm{n} $ and $ \bm{v} $. $ \bm{v} $ is the direction vector of the line and $ \bm{n} $ is the normal vector of the plane determined by the line and the origin \cite{sola2012impact}.

Since 3D line has only four degrees of freedom, the $ Pl\ddot{u}cker $ coordinates are over parameterized. In the back-end graph optimization, the extra degrees of freedom will increase the computational cost and cause the numerical instability of the system. 
Thus, Bartoli \cite{bartoli2005structure} proposed the orthonormal representation with minimum four parameters. We can obtain the orthonormal representation $ (\bm{U},\bm{W})\in{SO(3)}\times{SO(2)} $ from $ Pl\ddot{u}cker $ coordinates:

\begin{align}
	{\bm{\mathcal L}} &= \left[ {{\bm{n}}|{\bm{v}}} \right] = \left[ {\begin{array}{*{20}{c}}
	{\frac{{\bm{n}}}{{\left\| {\bm{n}} \right\|}}}&{\frac{{\bm{v}}}{{\left\| {\bm{v}} \right\|}}}&{\frac{{{\bm{n}} \times {\bm{v}}}}{{\left\| {{\bm{n}} \times {\bm{v}}} \right\|}}}
	\end{array}} \right]\left[ {\begin{array}{*{20}{c}}
	{\left\| {\bm{n}} \right\|}&0\\
	0&{\left\| {\bm{v}} \right\|}\\
	0&0
	\end{array}} \right]\\
 	&= {\bm{U}}\left[ {\begin{array}{*{20}{c}}
	{{w_1}}&0\\
	0&{{w_2}}\\
	0&0
	\end{array}} \right].
\end{align}
The orthonormal representation of line $ (\bm{U},\bm{W})$ consists of:
\begin{align}
	{\bm{U}} &= {\bm{U}(\bm{\theta})} = \left[ {\begin{array}{*{20}{c}}
			{{u_{11}}}&{{u_{12}}}&{{u_{13}}}\\
			{{u_{21}}}&{{u_{22}}}&{{u_{23}}}\\
			{{u_{31}}}&{{u_{32}}}&{{u_{33}}}
	\end{array}} \right]\\
{\bm{W}} &= {\bm{W}(\theta)}  = \left[ {\begin{array}{*{20}{c}}
			{{w_1}}&{ - {w_2}}\\
			{{w_2}}&{{w_1}}
	\end{array}} \right].
\end{align}

\begin{figure}[!h]
	\centering
	\includegraphics[width=.6\columnwidth]{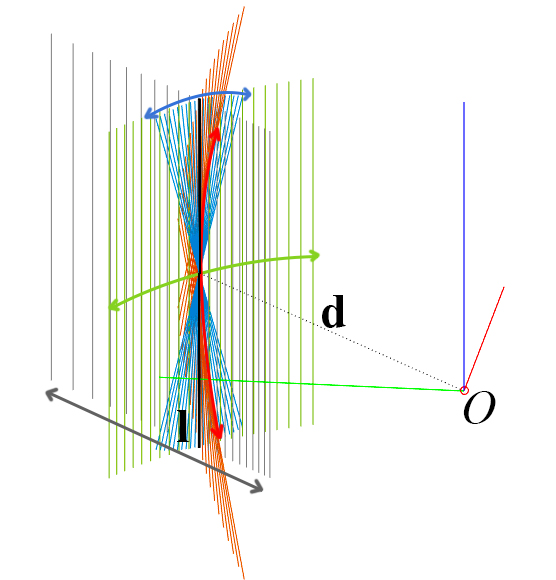}
	\caption{ Geometric interpretation of four parameters $ \bm{\delta}_{\theta}$ in updating orthonormal representation. }
	\label{3fig5}
\end{figure}

 We can update the orthonormal representation with the minimum four parameters
 $\bm{{\delta}}_{\theta}={[\bm{\theta}^T,\theta ]}^T\in{\mathbb{R}^4} $,
we can update $ \bm{U} $ with the vector $ \bm{\theta} \in{\mathbb{R}^3} $, and update $ \bm{W} $ with $ \theta $.
Each sub-parameter of $ \bm{\delta}_{\theta}$ has a specific geometric interpretation. $ \bm{W} $ updated by $ \theta $ encapsulates the vertical distance $ \bm{{\rm{d}}} $ from the origin to the spatial line. As shown in Fig. \ref{3fig5}, in the case of fixed $ \bm{W} $ represented in gray, the three-dimensional vector $ \bm{\theta} $ is related to the rotation of the line around three axes, drawn in orange, green, and blue.

Note that in the proposed visual SLAM system, 
we only use the orthonormal representation in the back-end optimization, as it is the minimal and decoupled representation. 
However, in the other steps, the $ Pl\ddot{u}cker $ coordinates are used due to its convenience in camera projection, endpoints trimming, and line initialization~\cite{bartoli2005structure,zhang2015building}.

%
%
%
%

\section{Graph Optimization with Point and Line Measurements}


In what follows, we present in detail how the line measurements are incorporated into our graph-based visual SLAM system,
while the point measurements are treated in a standard way, for example, as in ORB-SLAM~\cite{murORB2}.

\subsection{Measurement Models of Point and Line Features}\label{sec41}

We use the transformation matrix $ \bm{T}_{cw}\in{SE(3)} $ to denote the transformation from world frame to camera frame, which consists of a rotation matrix $ \bm{R}_{cw}\in{SO(3)} $, and a translation vector $ \bm{t}_{cw}\in{\mathbb{R}^3} $, as shown in~\eqref{eq4}. First, we convert the 3D line $ \bm{\mathcal{L}}_w $ from the world frame to the camera frame~\cite{bartoli20013d} as shown in \eqref{eq5}, denoted as $ \bm{\mathcal{L}}_c $, with representation of the $ Pl\ddot{u}cker $ coordinates. Then the 3D line $ \bm{\mathcal{L}}_c $ is projected to the image in~\eqref{eq6}, described as ${\bm{l}'}$ on image plane, according to the known intrinsic parameters of camera. It should be noted that only normal components $ \bm{n}_c $ in the $ Pl\ddot{u}cker $ coordinates $ \bm{\mathcal{L}}_c $ can provide meaningful information in the projection. Then, the re-projection error of 3D line is represented as the distance between two homogeneous endpoints $ \bm{x}_s $, $ \bm{x}_e $ of the matched line segment $ \bm{z} $ to the back-projected line ${\bm{l}'}$ on image plane as shown in~\eqref{eq7}. 

%
%
\begin{equation}
{\bm{T}_{cw}} = \left[ {\begin{array}{*{20}{c}}
	{{{\bm{R}}_{cw}}}&{{\bm{t}_{cw}}}\\
	\bm{0}&1
	\end{array}} \right]
\label{eq4}
\end{equation}
\begin{equation}
{\bm{\mathcal L}_c} = \left[ {\begin{array}{*{20}{c}}
	{{\bm{n}_c}}\\
	{{\bm{v}_c}}
	\end{array}} \right]={\bm{\mathcal{H}}_{cw}}{\bm{\mathcal L}_w} = \left[ {\begin{array}{*{20}{c}}
	{{{\bm{R}}_{cw}}}&{{{\left[ {{{\bm{t}}_{cw}}} \right]}_ \times }{{\bm{R}}_{cw}}}\\
	\bm{0}&{{{\bm{R}}_{cw}}}
	\end{array}} \right]{\bm{\mathcal L}_w},
\label{eq5}
\end{equation}
where ${\left[ . \right]_ \times }$ denotes the skew-symmetric matrix of a vector, and $ {\bm{\mathcal{H}}_{cw}} $ represents transformation matrix of the line.
\begin{equation}
{\bm{l}'}={\bm{\mathcal{K}}}{\bm{n}_c} = \left[ {\begin{array}{*{20}{c}}
	{{f_v}}&0&0\\
	0&{{f_u}}&0\\
	{ - {f_v}{c_u}}&{{f_u}{c_v}}&{{f_u}{f_v}}
	\end{array}} \right]{{\bm{n}}_c}=\left[ {\begin{array}{*{20}{c}}
	{{l_1}}\\
	{{l_2}}\\
	{{l_3}}
	\end{array}} \right],
\label{eq6}
\end{equation}
where $ \bm{\mathcal{K}} $ denotes the projection matrix of the line~\cite{sola2012impact}.
\begin{equation}
{e_l} = {\rm{d}}\left( {{\bm{z}},{\bm{l}'}} \right) = {\left[ {\frac{{{\bm{x}}_s^{\rm{T}}{\bm{l}'}}}{{\sqrt {l_1^2 + l_2^2} }},\frac{{{\bm{x}}_e^{\rm{T}}{\bm{l}'}}}{{\sqrt {l_1^2 + l_2^2} }}} \right]^{\rm{T}}},
\label{eq7}
\end{equation}
where $ {{\rm{d}}}(.) $ denotes the distance function.
%



The camera pose $ {\bm{T}}_{kw} $, the 3D point position $ {\bm{X}_{w,i}} $, and the position of 3D line $ {\bm{\mathcal{L}}_{w,j}} $ are denoted as vertices in the graph model. The two types of edge, the pose-point edge in~\eqref{geq8}, the pose-line edge in~\eqref{geq9}, are constructed according to the front-end data association. The re-projection errors encapsulated in the edges are:
\begin{equation}
{Ep_{k,i}}{{ = }}{{\bm{x}}_{k,i}} - \bm{{\rm{K}}}{\bm{T}_{kw}}{\bm{X}_{w,i}}
\label{geq8}
\end{equation}
\begin{equation}
{El_{k,j}} = {\rm{d}}\left( {{{\bm{z}}_{k,j}},\bm{\mathcal{K}}\rm{n}_c[{\bm{\mathcal{H}}_{cw}}{\bm{\mathcal{L}}_{w,j}}]} \right),
\label{geq9}
\end{equation}
where $ {{\bm{x}}_{k,i}} $ stands for the coordinates of point in the image, $\rm{n}_c[.]$ denotes the normal components of the $ Pl\ddot{u}cker $ coordinates. 
 For simplicity, we omit the conversion from homogeneous coordinates to the inhomogeneous in the above equations. Assuming that the observations obey Gaussian distribution, the final cost function $ C $ can be obtained as in \eqref{leq10}, Where $ \bm{\Sigma p}^{ - 1} $, $ \bm{\Sigma l}^{ - 1} $ are the inverse covariance matrices of points and lines, and $ {\rho _p} $, $ {\rho _l} $ are robust Huber cost functions. The back-end optimization minimizes the cost function $ C $.
\begin{equation}
\label{leq10}
C = \mathop \sum \limits_{k,i} {\rho _p}\left( {Ep_{k,i}^{\rm{T}}{\bm{\Sigma p}}_{k,i}^{ - 1}{Ep_{k,i}}} \right) + \mathop \sum \limits_{k,j} {\rho _l}\left( {El_{k,j}^{\rm{T}}{\bm{\Sigma l}}_{k,j}^{ - 1}{El_{k,j}}} \right)
\end{equation}

\subsection{Jacobian of Line Re-projection Error}	

It is known that the Jacobian is important when using an iterative approach to solve the the graph optimization problem. 
To the best of our knowledge, this is the first paper deriving out the analytical Jacobains of re-projection errors with respect to line parameters, which including the Jacobian with respect to the small pose changes ${{\bm{\delta }}_\xi }$, and to the four dimensional vector ${{\bm{\delta }}_\theta }$ which updates the orthonormal representation. The Jacobian of the line re-projection error ${el} = {{{\rm{d}}}}({\bm{z},{\bm{l}}'})$ with respect to the back-projected line ${\bm{l}'} = {[{l_1},{l_2},{l_3}]^T}$ is given by:
\begin{equation}
\frac{{\partial {el}}}{{\partial {\bm{l}}'}} 
= \frac{1}{{{l_n}}}{\left[ {\begin{array}{*{20}{c}}
		{{u_1} - \frac{{{l_1}{e_1}}}{{l_n^2}}}&{{v_1} - \frac{{{l_2}{e_1}}}{{l_n^2}}}&1\\
		{{u_2} - \frac{{{l_1}{e_2}}}{{l_n^2}}}&{{v_1} - \frac{{{l_2}{e_2}}}{{l_n^2}}}&1
		\end{array}} \right]_{2 \times 3}},
\end{equation}
where ${e_1} = {\bm{x}}_s^{\rm{T}}{\bm{l}}'$, ${e_2} = {\bm{x}}_e^{\rm{T}}{\bm{l}}'$, ${l_n} = \sqrt {(l_1^2 + l_2^2)} $. ${{\bm{x}}_s} = {\left[ {{u_1},{v_1},1} \right]^{\rm{T}}}$ and ${{\bm{x}}_e} = {\left[ {{u_2},{v_2},1} \right]^{\rm{T}}}$ are the two endpoints of matched line segment in the image.

Recall the projection of 3D line ${\bm{l}}' = \bm{\mathcal{K}}{{\bm{n}}_c}$, then:
\begin{equation}
\frac{{\partial {\bm{l}}'}}{{\partial {\bm{\mathcal{L}}_c}}} = \frac{{\partial {\cal K}{{\bm{n}}_c}}}{{\partial {\bm{\mathcal{L}}_c}}} = {\left[ {\begin{array}{*{20}{c}}
		\bm{\mathcal{K}}&{\bm{0}}
		\end{array}} \right]_{3 \times 6}}
\end{equation}
Assuming that the orthonormal representation of line in the world frame $ {\bm{\mathcal{L}} }_w$, which consists of $\bm{U}$ and $\bm{W}$,
	%
	We write the Jacobians directly: 
\begin{equation}
\frac{{\partial {{\bm{\mathcal{L}}}_c}}}{{\partial {{\bm{\mathcal{L}}}_w}}} = \frac{{\partial {{\bm{\mathcal{H}}}_{cw}}{{\bm{\mathcal{L}}}_w}}}{{\partial {{\bm{\mathcal{L}}}_w}}} = {{\bm{\mathcal{H}}}_{cw}}
\end{equation}
\begin{equation}
\frac{{\partial {{\bm{\mathcal{L}}}_w}}}{{\partial {{\bm{\delta }}_\theta }}} = {\left[ {\begin{array}{*{20}{c}}
		{ - {{\left[ {{w_1}{{\bm{u}}_1}} \right]}_ \times }}&{ - {w_2}{{\bm{u}}_1}}\\
		{ - {{\left[ {{w_2}{{\bm{u}}_2}} \right]}_ \times }}&{  {w_1}{{\bm{u}}_2}}
		\end{array}} \right]_{6 \times 4}},
\end{equation}
where $ \bm{u}_i $ is the $ i_{th} $ column of $ \bm{U} $.

It is difficult to compute $\frac{{\partial {{\bm{\mathcal{L}}}_w}}}{{\partial {{\bm{\delta }}_\xi }}}$ directly, so we divide the pose changes ${{\bm{\delta }}_\xi }$ into two parts, the translation part ${{\bm{\delta }}_\rho }$ and the rotation part ${{\bm{\delta }}_\phi }$. 
 ${{\bm{\delta }}_\phi }$ are set to zeros when computing Jacobian with respect to ${{\bm{\delta }}_\rho }$. With a transformation matrix ${\bm{T}^*}$ containing the translation ${{\bm{\delta }}_\rho }$, the new line ${\bm{\mathcal{L}}}_c^*$ is:


\begin{equation}
{\bm{T}^*} = \exp \left( {{{\bm{\delta }}_\xi} ^\wedge} \right){T_{{{cw}}}} \approx \left[ {\begin{array}{*{20}{c}}
	{\bm{I}}&{{{\bm{\delta }}_\rho }}\\
	{{{\bm 0}^{{T}}}}&1
	\end{array}} \right]{T_{{{cw}}}}
\end{equation}
\begin{equation}
{{\bm{R}}^*} = {{\bm{R}}_{cw}}\text{, }{{\bm{t}}^*} = {{\bm{\delta }}_\rho } + {{\bm{t}}_{cw}}
\end{equation}
\begin{equation}
{\bm{\mathcal{H}}}_{cw}^* = \left[ {\begin{array}{*{20}{c}}
	{{{\bm{R}}_{cw}}}&{{{\left[ {{{\bm{\delta }}_\rho } + {{\bm{t}}_{cw}}} \right]}_ \times }{{\bm{R}}_{cw}}}\\
	{\bm 0}&{{{\bm{R}}_{cw}}}
	\end{array}} \right]
\end{equation}
\begin{equation}
{\bm{\mathcal{L}}}_c^* = {\bm{\mathcal{H}}}_{cw}^*{{\bm{\mathcal{L}}}_w} = \left[ {\begin{array}{*{20}{c}}
	{{{\bm{R}}_{cw}}{{\bm{n}}_w} + {{\left[ {{{\bm{\delta }}_\rho } + {{\bm{t}}_{cw}}} \right]}_ \times }{{\bm{R}}_{cw}}{{\bm{v}}_w}}\\
	{{\bm{R}}_{cw}^{{T}}{{\bm{v}}_w}}
	\end{array}} \right],
\end{equation}
where $ \exp \left( {{{\bm{\delta }}_\xi} ^\wedge} \right) $ denotes the exponential map from Lie algebras to Lie Groups (hence $ {{{\bm{\delta }}_\xi} ^\wedge} $ is a Lie algebra). Then it is easy to deduce the partial derivative of $ \bm{\delta}_{\rho} $ :
\begin{equation}
\frac{{\partial {\bm{\mathcal{L}}}_c^*}}{{\partial {{\bm{\delta }}_\rho }}} = \left[ {\begin{array}{*{20}{c}}
	{\frac{{{{\left[ {{{\bm{\delta }}_\rho } + {{\bm{t}}_{cw}}} \right]}_ \times }{{\bm{R}}_{cw}}{{\bm{v}}_w}}}{{\partial {{\bm{\delta }}_\rho }}}}\\
	{\bm 0}
	\end{array}} \right] = {\left[ {\begin{array}{*{20}{c}}
		{ - {{\left[ {{{\bm{R}}_{cw}}{{\bm{v}}_w}} \right]}_ \times }}\\
		{\bm 0}
		\end{array}} \right]_{6 \times 3}}
\end{equation}
The process to deduce $\frac{{\partial {\bm{\mathcal{L}}}_c^*}}{{\partial {{\bm{\delta }}_\phi }}}$ is similar to $\frac{{\partial {\bm{\mathcal{L}}}_c^*}}{{\partial {{\bm{\delta }}_\rho }}}$, except for ${{\bm{\delta }}_\rho } = {\bm 0}$. We only shows the final result Eq.\ref{4eq20}, and drops the coordinate frame subscripts for readability. Readers can refer to the Appendix for more details.

\begin{equation}
\frac{{\partial {\bm{\mathcal{L}}}_c^*}}{{\partial {{\bm{\delta }}_\phi }}} = = {\left[ {\begin{array}{*{20}{c}}
		{ - {{\left[ {{\bm{Rn}}} \right]}_ \times } - {{\left[ {{{\left[ {\bm{t}} \right]}_ \times }{\bm{Rv}}} \right]}_ \times }\;}\\
		{ - {{\left[ {{\bm{Rv}}} \right]}_ \times }}
		\end{array}} \right]_{6 \times 3}}
	\label{4eq20}
\end{equation}
Stacking the Jacobians of ${{\bm{\delta }}_\rho }$ and ${{\bm{\delta }}_\phi }$ , we can obtain the final Jacobian of ${{\bm{\delta }}_\xi }$:
%
\begin{equation}
 \frac{{\partial {\bm{\mathcal{L}}}_c^*}}{{\partial {{\bm{\delta }}_\xi }}} = {\left[ {\begin{array}{*{20}{c}}
		{ - {{\left[ {{{\bm{R}}}{{\bm{n}}}} \right]}_ \times } - {{\left[ {{{\left[ {{{\bm{t}}}} \right]}_ \times }{{\bm{R}}}{{\bm{v}}}} \right]}_ \times }}&{ - {{\left[ {{{\bm{R}}}{{\bm{v}}}} \right]}_ \times }}\\
		{ - {{\left[ {{{\bm{R}}}{{\bm{v}}}} \right]}_ \times }}&{\bm 0}
		\end{array}} \right]_{6 \times 6}}
\end{equation}

Finally, the Jacobian of the re-projection error with respect to the line parameters can be found using the chain rule:
\begin{equation}
J{l_\xi } = \frac{{\partial {e_l}}}{{\partial {{\bm{\delta }}_\xi }}} = \frac{{\partial {e_l}}}{{\partial {\bm{l}}'}}\frac{{\partial {\bm{l}}'}}{{\partial {{\bm{\mathcal{L}}}_c}}}\frac{{\partial {{\bm{\mathcal{L}}}_c}}}{{\partial {{\bm{\delta }}_\xi }}}
\end{equation}
\begin{equation}
J{l_\theta } = \frac{{\partial {e_l}}}{{\partial {{\bm{\delta }}_\theta }}} = \frac{{\partial {e_l}}}{{\partial {\bm{l'}}}}\frac{{\partial {\bm{l'}}}}{{\partial {{\bm{\mathcal{L}}}_c}}}\frac{{\partial {{\bm{\mathcal{L}}}_c}}}{{\partial {{\bm{\mathcal{L}}}_w}}}\frac{{\partial {{\bm{\mathcal{L}}}_w}}}{{\partial {{\bm{\delta }}_\theta }}}
\end{equation}
Once these analytical Jacobians are available, we can employ iterative algorithms such as Gaussian-Newton to solve the graph optimization problem.

\section{Experimental Results}

%
\subsection{System Implementation}

\begin{figure}[!h]
	\centering
	\includegraphics[width=\columnwidth]{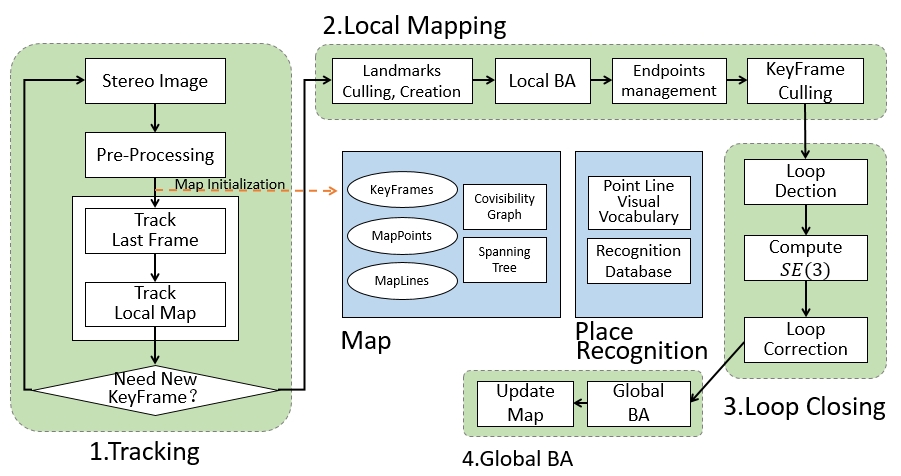}
	\caption{The architecture of the proposed graph-based visual SLAM system using both point and line features.}
	\label{pfig1}
\end{figure}

The proposed visual SLAM system is designed and implemented based on ORB-SLAM2~\cite{murORB2} and has three main parallel threads (see Fig. \ref{pfig1}): Tracking, Local Mapping and Loop Closing. 
The global BA thread is started only after finishing loop closing. 
In the following, we briefly describe each component while focusing on the difference from~\cite{murORB2}.

\subsubsection{Tracking}

\begin{figure}[!h]
	\centering
	\includegraphics[width=.85\columnwidth]{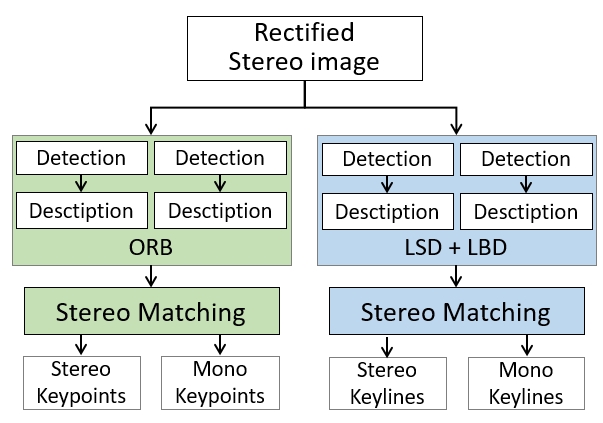}
	\caption{The workflow of pre-processing images.}
	\label{pfig2}
\end{figure}

Our system uses rectified stereo image sequence as input. For every input frame, four threads are launched to extract point feature (Keypoints) and line feature (Keylines) for left and right image in parallel. ORB features are applied for point feature detection and description. Line feature is detected by LSD and described by LBD descriptor. Then two threads are launched for stereo matching and all the features are classified as stereo or monocular features according to whether the feature in the left image could find its stereo match in the right image, as shown in Fig. \ref{pfig2}. The stereo matching of 3D lines performs as described in Section~\ref{seclinematch}. For each monocular feature, we search a match with other unmatched features in other keyframes. Once finding the matched feature, we triangulate the feature in the same way as stereo features. 
%
%
%
%

Motion estimation is made by two types of tracking, namely tracking last frame and tracking local map. The former one gives an initial pose estimation with the correspondences of adjacent frame, while the latter one refines the pose with much more constraints between the current frame and the local map. After the data association, the pose is estimated by motion-only BA using Levenberg-Marquardt algorithm~\cite{more1978levenberg}.

\begin{figure}[!h]
	\centering
	\includegraphics[width=.5\columnwidth]{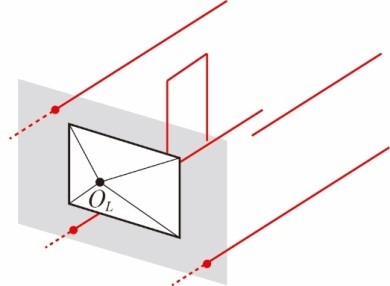}
	\caption{ Partial observation of 3D line. (The dash lines can't be observed by the camera while the solid lines can. The red points denotes the intersection of the plane and the 3D line.)}
	\label{pfig3}
\end{figure}

We use a constant velocity motion model to predict the camera pose as a prior when tracking last frame. Once the prior is known, the map points and the map lines in the last frame or the local map can be projected to current frame to build more associations. Then we perform a guided search to bound the complexity and enhance the accuracy of matching. Since the 3D line may be partially observed, the projected 2D line cannot be handled the same as the projected 2D point. Fig. \ref{pfig3} shows a simple example, the dash lines can't be observed by the camera while the solid lines can. In order to ensure the visibility of the projected 2D line segments in image plane, we propose a culling based method described as follow:
\begin{enumerate}
	\item Transform the 3D line $ \bm{\mathcal{L}}_w $ from world frame to current frame according to the prior $ {{\bm T}_{kw}}' $. Compute the two endpoints $ \bm{X}_{sk} $ and $ \bm{X}_{ek} $.
	\item Discard the line if both $ \bm{X}_{sk} $ and $ \bm{X}_{ek} $ are behind the camera. If one of them is behind the camera, compute the intersection of the plane and the 3D line by ${{\bm{X}}_{ik}} = {{\bm{X}}_{sk}} + \lambda \left( {{{\bm{X}}_{sk}} - {{\bm{X}}_{ek}}} \right)\text{ , where $ \lambda $ is a value between 0 and 1}$, as depicted in Fig. \ref{pfig3}.
	\item Project the two 3D endpoints in front of the camera to image plane. Since the projected line maybe lays across or even out of the image bound, all the projected lines must be dealt by Liang-Barsky algorithm \cite{liang1984new} which is an efficient line clipping algorithm and can retain the orientation of original line.
\end{enumerate}
Then line matching can be done efficiently thanks to the restricted searching space and binary descriptor. The last step is to decide whether the current frame is a new keyframe. We adopt the same policy as ORB-SLAM2~\cite{murORB2} and add more conditions related to line features.

\subsubsection{Local Mapping}

Once a new keyframe is added, the connections between the current keyframe and other frames will be updated by providing the co-visible information. Local mapping triangulates more map points and lines, removes outlier landmarks, and deletes redundant keyframe. All the camera poses and landmarks in the local map are adjusted by performing local BA. During back-end optimization, the 3D line is parameterized by infinite spatial line, hence its endpoints have no affect on the final optimization results. However, the endpoints play an important role in matching and visualizing, so our system need to maintain two endpoints of the 3D line after optimization. It can be done by back-projecting the 2D line in current keyframe and trimming the corresponding 3D line, which is similar to SLSLAM~\cite{zhang2015building}.

\subsubsection{Loop Closing and Global BA}

The loop closing thread is used to reduce drift accumulated during exploration by loop detection and loop correction. Loop detection try to find candidate keyframes based on the technique of bags of words. The visual vocabulary should be trained offline with both point and line features. Here we cluster the ORB features and LBD features to build their own vocabulary by DBOW~\cite{galvez2012bags} respectively. For every input keyframe, it is converted to the bag of words vector and stored in the online database. The similarity score between two bag of vector $ \bm{v}_a $ and $ \bm{v}_b $ can be computed as follow:
\begin{equation}
{\rm{s}} = {{\lambda \rm{s}_p}}{\left( {{{\bm{v}}_{{a}}},{{\bm{v}}_{{b}}}} \right)} + \left( {1 - {{\lambda }}} \right){\rm{s}_l}{\left( {{{\bm{v}}_{{a}}},{{\bm{v}}_{{b}}}} \right)},
\end{equation}
where $ \lambda $ is an empirical weight coefficient related to scenes. ${\rm{s}_p}{\left( {{{\bm{v}}_{{a}}},{{\bm{v}}_{{b}}}} \right)}$ and ${\rm{s}_l}{\left( {{{\bm{v}}_{{a}}},{{\bm{v}}_{{b}}}} \right)}$ are the similarity score of point feature and line feature. Then we can find the correspondences between the new keyframe and the candidate keyframe. we also refine the correspondences with time consistency test~\cite{mur2014fast}. And try to compute a $ SE(3) $ transformation matrix by EPnP \cite{lepetit2009epnp} with corresponding points in a RANSAC scheme \cite{fischler1981random}. If failed, we alternatively compute a $ SE(3) $ by a method proposed in \cite{pradeep2012egomotion} using the matching lines across two stereo views. Finally, a pose graph optimization is launched to correct the loop. Once finished, a global BA is incorporated to achieve the optimal solution in a separate thread.

\subsection{Results}

Various experiments have been conducted in both synthetic and real-world scene. The accuracy and time efficiency of our approach are analyzed. In these experiments, the algorithm run on a computer with Intel Core i7-2600 @ 3.40GHz and 16GB memory in a 64-bit Linux operating system. 

\subsubsection{Synthetic data}

  \begin{figure}[!h]
	\centering
	\includegraphics[width=\columnwidth]{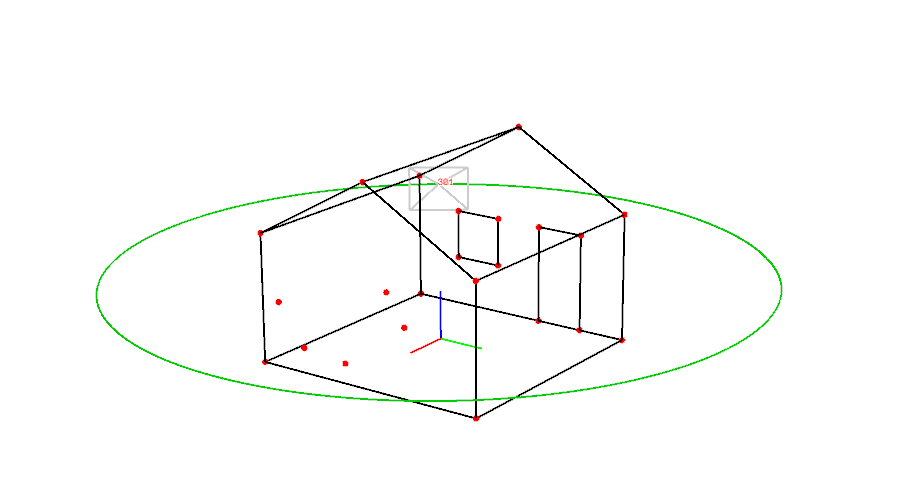}
	\caption{ Synthetic scene with 25 lines and variable number of points. }
	\label{efig1}
\end{figure}
\begin{figure}[!h]
	\centering
	\includegraphics[width=\columnwidth]{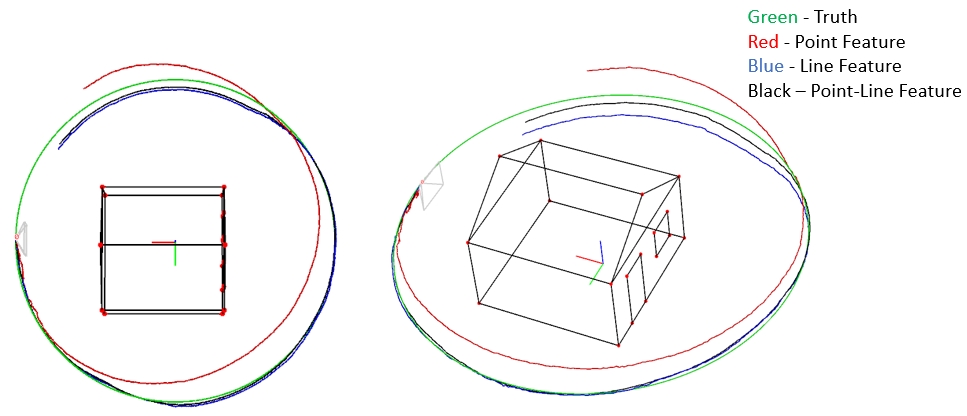}
	\caption{Top and oblique views of estimated camera trajectory. }
	\label{efig2}
\end{figure}

There is an accurate data association in synthetic Scene. And this experiment is proposed to verify the correctness and advantage of the introduced line feature in the back-end optimization. The derived Jacobian of 3D line re-projection error is used in the optimization. The synthetic scene in Fig. \ref{efig1} contains a house with a total of 25 lines and variable number of points. This construction is similar to the scene in \cite{sola2012impact}. Virtual stereo camera with baseline of 0.5m moves around the house, collecting images of $ 640\times480 $ pixels. Gaussian white noise with a variance of 1 pixel are added to the points and endpoints of lines in the captured images. Loop detection is disabled to display pose error clearly. $ RMSE $ (Root Mean Square Error) of $ RPE $ (Relative Pose Error) is the metric to evaluate the performance of our method. Fig. \ref{efig2} shows an estimated trajectory by our proposed system. The average result of Monte Carlo experiments of 25 runs, is shown in Table \ref{etab1}. $ RPEtrans1 $ and $ RPErot1 $ are translation and rotation errors obtained in the scene with lots of point features, while $ RPEtrans2 $ and $ RPErot2 $ result from the scene containing few points. In the scene with comparable number of points and lines, odometry based on only point feature performs better than one using only lines. The reason may be that re-projection error of an infinite long spatial line is only related to the normal vector of the $ Pl\ddot{u}cker $ line coordinates as shown in Section~\ref{sec41}. So the matched point features produce more constrains than the same number lines. The table shows that the method based on point features has a larger error than on line features in the scene with few points. Our method based on fusion of points and lines outperform than the both.

\begin{table}[!h]
	\caption{RPE of the Methods Based on Different Features} \vspace{-2em}
	\label{etab1}
	\begin{center}
		\begin{tabular}{l|c|c|c}\hline
			&Point Feature &Line Feature& Point-Line Feature \\
			\hline
			$ RPEtrans1(m) $ & 0.08702&	0.09827&	\textbf{0.07852}\\
			\hline
			$ RPErot1(rad) $ & 0.00430&	0.00486&	\textbf{0.00381}\\
			\hline
			$ RPEtrans2(m) $ & 0.19254&	0.09621&	\textbf{0.08637}\\
			\hline
			$ RPErot2(rad) $ & 0.00798&	0.00481&	\textbf{0.00408}\\
			\hline
		\end{tabular}
	\end{center}
\end{table}

\subsubsection{Real data}

The real-world scene experiment is carried on both it3f dataset \cite{zhang2015building} and KITTI dataset \cite{geiger2013vision}. For a more comprehensive assessment of the approach presented in this article, several open source approaches are compared in this section, including ORB-SLAM2 \cite{murORB2}, SLSLAM \cite{zhang2015building}, PLSVO \cite{gomez2016robust} and PL-SLAM presented in this paper. ORB-SLAM2 is a complete point feature based SLAM system that contains map reuse, loop detection and relocation. SLSLAM is based on the straight line feature, constructing scene composed of only straight lines, which is a relatively excellent line based SLAM system. PLSVO is only an odometry using two endpoints to represent the spatial line and performing brute-force matching in the front-end.

  \begin{figure}[!h]
	\centering
	\includegraphics[width=.9\columnwidth]{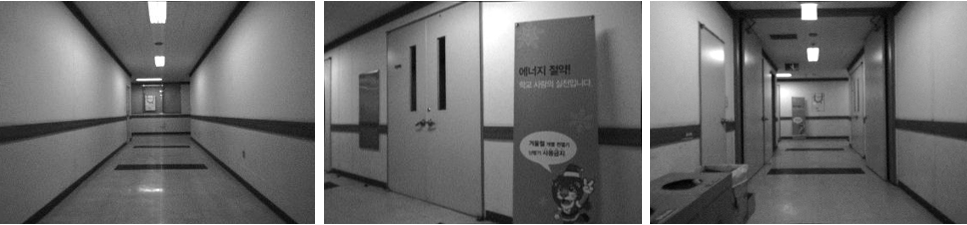}
	\caption{ Sample images used in it3f dataset~\cite{zhang2015building}. }
	\label{efig3}
\end{figure}

  \begin{figure}
	\centering
	\includegraphics[width=0.45\textwidth]{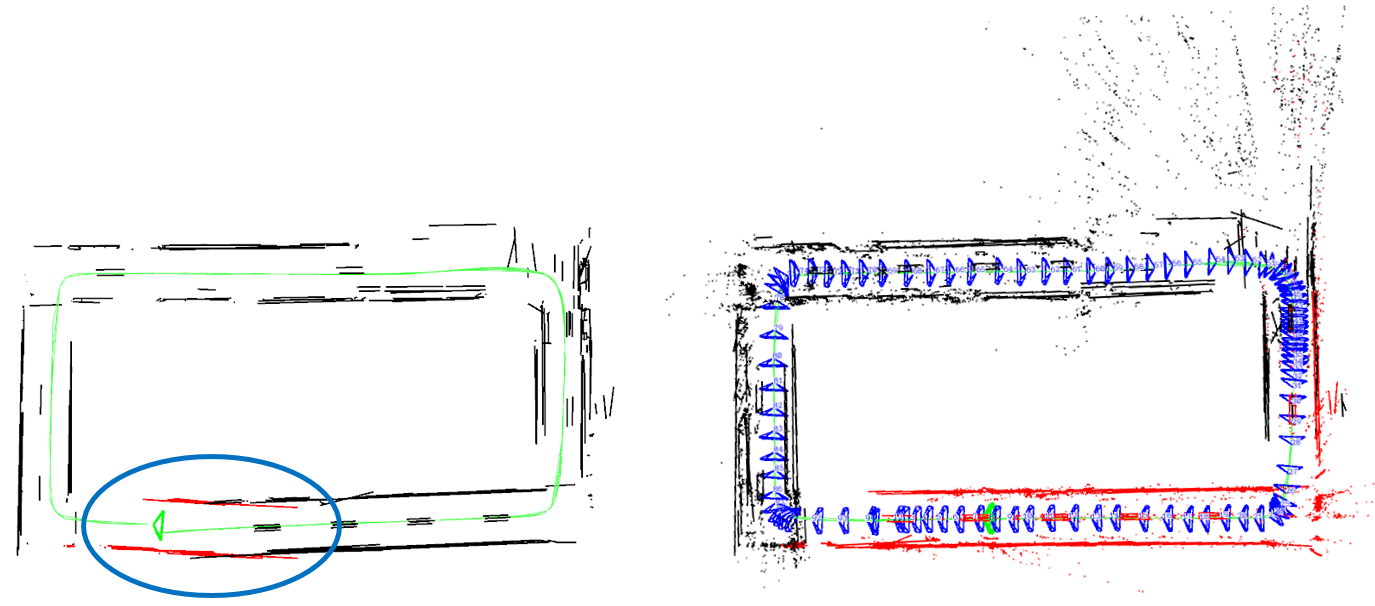}
	\caption{ Results before and after of the loop closure. Left: Results before the loop closure. Right: Results after the loop closure and loop correction. }
	\label{efig5}
\end{figure}

Fig. \ref{efig3} shows images from the it3f dataset, and Fig. \ref{ifig1} shows the results generated from this dataset. Fig. \ref{efig5} shows the trajectory and map of the camera before and after a loop closure followed by a bundle adjustment. PLSVO has a poor performance on this dataset, so we only compare ORB-SLAM2 and SLSLAM with our proposed PL-SLAM. it3f dataset doesn't provided the ground truth. The degrees of drift before the loop closure are compared. For fair comparison, both ORB-SLAM2 and PL-SLAM disable the loop detection thread, and use the same parameters in point feature extraction. For each image, we extract 1000 point features at 8 scale levels with a scale factor of 1.2. 


\begin{table}[h]
	\caption{Errors Before Loop Closure} \vspace{-1em}
	\label{etab2}
	\begin{center}
		\begin{tabular}{c|c}\hline
			Method& Errors Before Loop Closure\\
			\hline
			PL-SLAM & $ {[-0.3549,1.4056,-0.0273]}^T $\\
			ORB-SLAM2 & $ {[-0.3748,1.9065,0.17442]}^T $\\
			SLSLAM & $ {[-0.3141,-0.5455,-0.06449]}^T $\\
			\hline
		\end{tabular}
	\end{center}
\end{table}

\begin{table*}[!h]
	\caption{Results of ORB-SLAM, PLSVO and PL-SLAM on KITTI Dataset} \vspace{-1em}
	\label{etab3}
	\begin{center}
		\begin{tabular}{c|c|c|c|c|c|c|c|c|c}\cline{2-10}
			&\multicolumn{3}{c|}{PLSVO}&\multicolumn{3}{c|}{ORB-SLAM2}&\multicolumn{3}{c}{PL-SLAM}\\\cline{2-10}
			&Trans(m)&Rot(rad)&ATE(m)&Trans(m)&Rot(rad)&ATE(m)&Trans(m)&Rot(rad)&ATE(m)\\\hline 
			sequence 03&	0.2247&	0.0046&	13.2415&	0.1598&	0.0023&	2.6638&	0.1601&	0.0024&	\textbf{2.6203}\\
			sequence 04&	0.2045&	0.0019&	2.3020&	0.1180&	0.0015&	0.7415&	0.1183&	0.0017&	\textbf{0.3663}\\
			sequence 10&	0.1809&	0.0053&	9.0208&	0.1143&	0.0022&	6.3974&	0.1166&	0.0021&	\textbf{5.9207}\\\hline
			
		\end{tabular}
	\end{center}
\end{table*}

\begin{figure}[!h]
	\centering
	\includegraphics[width=\columnwidth]{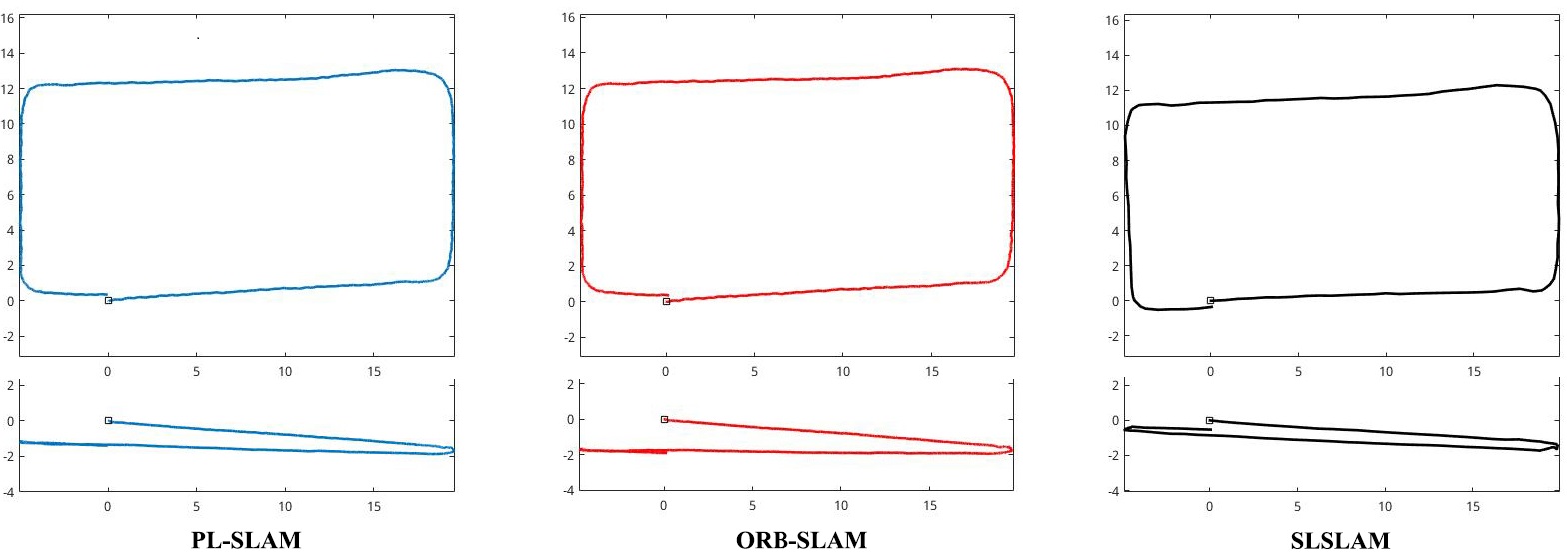}
	\caption{ Comparison results on it3f dataset~\cite{zhang2015building} without loop closure. The top and bottom row show the top and side views of the results. }
	\label{efig6}
\end{figure}

 Fig. \ref{efig6} shows the top and side views of the reconstruction results by the three systems without loop closures, respectively. The point with zero coordinates is the starting point and the other is the finishing point. Table \ref{etab2} shows the drift before the loop closure (translation in $ X(m), Y(m) , Z(m) $). It can be observed from the table that PL-SLAM perform better than ORB-SLAM2, which demonstrate the strength of including constraints of straight line. SLSLAM has the best performance, only -0.5455 meters error in the vertical direction. A reason can account for this is that it3f dataset contains low-textured scenarios, reflective white walls, windows and floor etc. At the same time, due to the influence of the ceiling lights, point features prone to be mismatched and bring big errors. In the optimization process of our proposed approach, we don't set different weights to the error terms of points and lines in \eqref{leq10} with consideration of versatility. When the component based on point feature has unstable performance and low accuracy, the proposed system based on combination of point and line features will be affected, which coincides with the synthetic scene experiment.

In terms of time efficiency, the execution time will not increase much because features are extracted in parallel threads. For images with dimensions of $ 640\times480 $, the feature extraction and stereo matching in ORB-SLAM2 requires 32.15ms, while our system requires 42.401ms with additional consideration of line features on it3f dataset. Our tracking thread can achieve a performance of 15.1 frame/s, which can satisfy the real-time requirements.

  \begin{figure}[!h]
 	\centering
 	\includegraphics[width=\columnwidth]{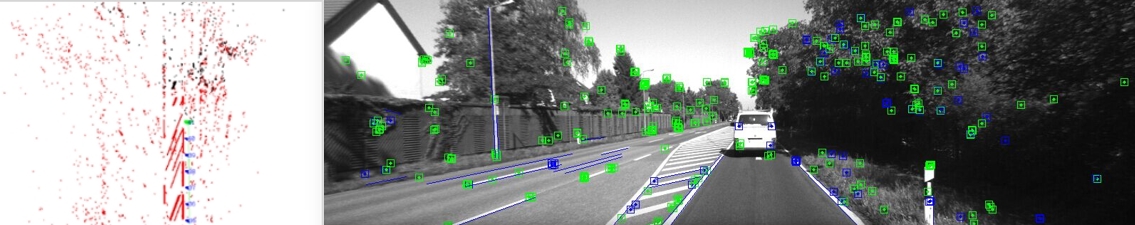}
 	\caption{ Results on KITTI dataset. Left: The map composed of point and line features. 
	Right: One frame with extracted point and line features. }
 	\label{efig7}
 \end{figure}

 We also evaluate our system on KITTI odometry benchmark. Sequences 03, 04, and 10, which have scenarios with lines, are selected. Fig. \ref{efig7} shows the result on KITTI dataset in our system. In this experiment, we only compare our PL-SLAM with ORB-SLAM2 and PLSVO\footnote{As the source code of the front-end module in SLSLAM is unavailable, we do not include it in the experiments on KITTI dataset.}. Loop detection modules are all disabled for fair comparison. 
%
%
%
%
In this experiment, $ RPE $ and $ ATE $ (Absolute Trajectory Error) are used as evaluation criterion. Table \ref{etab3} shows the results of the experiment, where $ Trans $ and $ Rot $ represent $ RPE $ of the translations and rotations respectively. The smallest $ ATE $ in each sequence is marked in the table.
%
%
%
%
%
It is shown that our system has acceptable performance in several sequences. A performance improvement can be achieved compared to the original ORB-SLAM2. PLSVO has a poor performance because of the brute-force matching in data association and accumulated errors. 




\section{Conclusions}

To improve the accuracy and robustness of visual SLAM, we present a graph-based approach using point and line features. The spatial line is expressed by the orthonormal representation in the optimization process, which is the compactest and decoupled form. And the Jacobians of re-projection error with respect to the line parameters is also derived out to make a good performance.
It is proved that fusing these two types of features will produce more robust estimation in synthetic and real-world scene. Our robust visual SLAM is also able to work in real-time. In the future, we will investigate how to introduce inertial sensors into our system with point and line features.




\section*{Appendix}
This appendix will explain the Jacobian with respect to ${{\bm{\delta }}_\phi }$ in detail. ${{\bm{\delta }}_\rho }$ are set to zeros when computing Jacobian with respect to ${{\bm{\delta }}_\phi }$. With a transformation ${{\bm{T}}^*}$ containing rotation ${{\bm{\delta }}_\phi }$, the new 3D line is denoted as ${\bm{\mathcal{L}}}_c^*$:

\begin{equation}
{{\bm{T}}^*} = \exp \left( {{{\bm{\delta }}_\xi} ^\wedge} \right){{\bm{T}}} \approx \left( {{\bm{I}} + \left[ {\begin{array}{*{20}{c}}
		{{{\left[ {{{\bm{\delta }}_\phi }} \right]}_ \times }}&{\bm 0}\\
		{{{\bm 0}^{\bm{T}}}}&{0}
		\end{array}} \right]} \right){{\bm{T}}}
\end{equation}
\begin{equation}
{{\bm{R}}^*} = \left( {{\bm{I}} + {{\left[ {{{\bm{\delta }}_\phi }} \right]}_ \times }} \right){{\bm{R}}}\text{, }{{\bm{t}}^*} = \left( {{\bm{I}} + {{\left[ {{{\bm{\delta }}_\phi }} \right]}_ \times }} \right){{\bm{t}}}
\end{equation}
\begin{equation}
{\bm{\mathcal{H}}}_{cw}^* = \left[ {\begin{array}{*{20}{c}}
	{\left( {{\bm{I}} + {{\left[ {{{\bm{\delta }}_\phi }} \right]}_ \times }} \right){\bm{R}}}&{\left( {{\bm{I}} + {{\left[ {{{\bm{\delta }}_\phi }} \right]}_ \times }} \right){{\left[ {\bm{t}} \right]}_ \times }{\bm{R}}}\\
	{\bm 0}&{\left( {{\bm{I}} + {{\left[ {{{\bm{\delta }}_\phi }} \right]}_ \times }} \right){\bm{R}}}
	\end{array}} \right]
\end{equation}
\begin{small}
\begin{equation}
{\bm{\mathcal{L}}}_c^* = {\bm{\mathcal{H}}}_{cw}^*{{\bm{\mathcal{L}}}_w} = \left[ {\begin{array}{*{20}{c}}
	{\left( {{\bm{I}} + {{\left[ {{{\bm{\delta }}_\phi }} \right]}_ \times }} \right){\bm{Rn}} + \left( {{\bm{I}} + {{\left[ {{{\bm{\delta }}_\phi }} \right]}_ \times }} \right){{\left[ {\bm{t}} \right]}_ \times }{\bm{Rv}}}\\
	{\left( {{\bm{I}} + {{\left[ {{{\bm{\delta }}_\phi }} \right]}_ \times }} \right){\bm{Rv}}}
	\end{array}} \right],
\end{equation}
\end{small}
where ${\left[ . \right]_ \times }$ denotes the skew-symmetric matrix of a vector. In the process of deducing ${\bm{\mathcal{H}}}_{cw}^*$, the properties of rotation matrix $\left( {{\bm{Ra}}} \right) \times \left( {{\bm{Rb}}} \right) = {\bm{R}}\left( {{\bm{a}} \times {\bm{b}}} \right)\text{, }{\bm{R}} \in SO\left( 3 \right)$ is used.
Then $\frac{{\partial {\bm{\mathcal{L}}}_c^*}}{{\partial {{\bm{\delta }}_\phi }}}$ can be written directly:
\begin{align}
\frac{{\partial {\bm{\mathcal{L}}}_c^*}}{{\partial {{\bm{\delta }}_\phi }}} &= \left[ {\begin{array}{*{20}{c}}
	{\frac{{\partial {{\left[ {{{\bm{\delta }}_\phi }} \right]}_ \times }{\bm{Rn}}}}{{\partial {{\bm{\delta }}_\phi }}} + \frac{{\partial {{\left[ {{{\bm{\delta }}_\phi }} \right]}_ \times }{{\left[ {\bm{t}} \right]}_ \times }{\bm{Rv}}}}{{\partial {{\bm{\delta }}_\phi }}}}\\
	{\frac{{\partial {{\left[ {{{\bm{\delta }}_\phi }} \right]}_ \times }{\bm{Rv}}}}{{\partial {{\bm{\delta }}_\phi }}}}
	\end{array}} \right]\\
&= \left[ {\begin{array}{*{20}{c}}
	{ - \frac{{{{\left[ {{\bm{Rn}}} \right]}_ \times }{{\bm{\delta }}_\phi }}}{{\partial {{\bm{\delta }}_\phi }}} - \frac{{\partial {{\left[ {{{\left[ {\bm{t}} \right]}_ \times }{\bm{Rv}}} \right]}_ \times }{{\bm{\delta }}_\phi }}}{{\partial {{\bm{\delta }}_\phi }}}}\\
	{ - \frac{{{{\left[ {{\bm{Rv}}} \right]}_ \times }{{\bm{\delta }}_\phi }}}{{\partial {{\bm{\delta }}_\phi }}}}
	\end{array}} \right]\\
& = {\left[ {\begin{array}{*{20}{c}}
		{ - {{\left[ {{\bm{Rn}}} \right]}_ \times } - {{\left[ {{{\left[ {\bm{t}} \right]}_ \times }{\bm{Rv}}} \right]}_ \times }\;}\\
		{ - {{\left[ {{\bm{Rv}}} \right]}_ \times }}
		\end{array}} \right]_{6 \times 3}}
\end{align}

%
%

\addtolength{\topmargin}{0.249cm}


%
\bibliographystyle{ieeetr}
\bibliography{mycitebib}

\end{document}